\documentclass[letterpaper]{article} 
\usepackage{aaai2026}  
\usepackage{times}  
\usepackage{helvet}  
\usepackage{courier}  
\usepackage[hyphens]{url}  
\usepackage{graphicx} 
\usepackage{amssymb}
\usepackage{amsmath}
\usepackage{multirow}
\usepackage{booktabs}
\usepackage{array}
\usepackage{makecell}
\usepackage{natbib}  
\usepackage{caption} 
\usepackage{algorithm}
\usepackage{algorithmic}

\definecolor{clcl}{RGB}{255,0,0}

\usepackage{newfloat}
\usepackage{listings}
\DeclareCaptionStyle{ruled}{labelfont=normalfont,labelsep=colon,strut=off} 
\floatstyle{ruled}
\newfloat{listing}{tb}{lst}{}
\floatname{listing}{Listing}
\nocopyright
\title{MindShot: Multi-Shot Video Reconstruction from fMRI with LLM Decoding}
\author{
    Wenwen Zeng\textsuperscript{\rm 1}\equalcontrib,
    Yonghuang Wu\textsuperscript{\rm 1}\equalcontrib,
    Yifan Chen\textsuperscript{\rm 1},
    Xuan Xie\textsuperscript{\rm 1},
    Chengqian Zhao\textsuperscript{\rm 1},
    Feiyu Yin\textsuperscript{\rm 1},
    Guoqing Wu\textsuperscript{\rm 1},
    Jinhua Yu\textsuperscript{\rm 1}\thanks{Correponding Author.}\\
}
\affiliations{
    \textsuperscript{\rm 1}Fudan University\\

}

\usepackage{bibentry}

\begin{document}

\maketitle

\begin{abstract}
Reconstructing dynamic videos from fMRI is important for understanding visual cognition and enabling vivid brain-computer interfaces. However, current methods are critically limited to single-shot clips, failing to address the multi-shot nature of real-world experiences. Multi-shot reconstruction faces fundamental challenges: fMRI signal mixing across shots, the temporal resolution mismatch between fMRI and video obscuring rapid scene changes, and the lack of dedicated multi-shot fMRI-video datasets. To overcome these limitations, we propose a novel divide-and-decode framework for multi-shot fMRI video reconstruction. Our core innovations are: (1) A shot boundary predictor module explicitly decomposing mixed fMRI signals into shot-specific segments. (2) Generative keyframe captioning using LLMs, which decodes robust textual descriptions from each segment, overcoming temporal blur by leveraging high-level semantics. (3) Novel large-scale data synthesis (20k samples) from existing datasets. Experimental results demonstrate our framework outperforms state-of-the-art methods in multi-shot reconstruction fidelity. Ablation studies confirm the critical role of fMRI decomposition and semantic captioning, with decomposition significantly improving decoded caption CLIP similarity by 71.8$\%$. This work establishes a new paradigm for multi-shot fMRI reconstruction, enabling accurate recovery of complex visual narratives through explicit decomposition and semantic prompting.
\end{abstract}

\section{Introduction}
Functional magnetic resonance imaging (fMRI) is a powerful, non-invasive tool for studying the human brain, particularly the visual system, through indirect measurement of neural activity \cite{horikawa2017generic}. Reconstructing dynamic visual sequences from fMRI data is critical not only for advancing our understanding of dynamic visual perception and cognition, but also for developing next-generation brain-computer interfaces (BCIs) capable of more vivid and dynamic “mind-reading” applications \cite{wen2018neural,fang2020reconstructing,fang2023extracting}. However, existing video reconstruction research mainly focuses on short-duration, single-shot videos (depicting a single, continuous scene or event) \cite{sun2025neuralflix,chen2023cinematic,li2024enhancing,luanimate}, ignoring the multi-shot visual experiences that characterize real-world cognition, such as watching films or recalling episodic memories.

Reconstructing multi-shot video presents substantial challenges beyond single-shot reconstruction. A primary limitation is the temporal mixing of neural signals corresponding to different shots. Existing methods often aggregate fMRI signals across entire videos into a single representation. This approach inherently merges the distinct neural activities associated with each individual shot, making it difficult to disentangle and accurately reconstruct the separate visual events. Furthermore, the inherent temporal resolution mismatch between fMRI and video obscures rapid visual changes like object motion and scene transitions, as hemodynamic responses integrate neural activity over seconds while visual events unfold at millisecond scales. This fundamentally undermines precise semantic alignment across shots, yielding semantically imprecise reconstructions. Moreover, the scarcity of large-scale fMRI datasets specifically designed for multi-shot video reconstruction further constrains the development of effective encoding and decoding strategies.


To address these challenges, we propose a novel divide-and-decode framework for multi-shot video reconstruction. Inspired by text-to-video generation paradigms where a narrative is segmented into prompts corresponding to individual shots \cite{zhao2024moviedreamer,wu2025automated}, we introduce a shot boundary predictor for fMRI segmentation. Instead of aggregating the entire fMRI data, this shot boundary predictor learns to segment it into shot-specific components corresponding to individual shots, enabling explicit and independent reconstruction of each shot. To address the limitations imposed by the fMRI-video temporal resolution mismatch, we propose to decode keyframe captions from fMRI data to achieve semantically precise reconstruction. For each segmented shot-specific fMRI, we decode a textual caption describing the keyframe using Large Language Models (LLMs). This leverages the observation that humans remember salient events at a semantic level, which is more robust to temporal blurring. The decoded caption then provides a precise semantic prompt for the subsequent video generation stage. To overcome data scarcity, we develop novel synthesis strategies to construct a large-scale multi-shot fMRI-video dataset. Leveraging existing publicly available fMRI-video datasets, including the benchmark CC2017 \cite{wen2018neural} and the dataset by Li et al. \cite{chen2023cinematic}, we synthesize 20k sample pairs for each dataset, enabling effective training of our proposed model.

Our contributions in this work can be summarized as follows:
\begin{itemize}
\item We propose a shot boundary predictor for shot-specific fMRI segmentation, enabling the decomposition of mixed fMRI signals into shot-specific components for multi-shot video reconstruction.
\item We propose a new semantics prediction method that decodes keyframe captions from shot-specific fMRI signals, mitigating semantic ambiguity caused by the fMRI-video temporal resolution mismatch and providing precise prompts for video generation.
\item We develop novel synthesis strategies to create large-scale multi-shot training data from existing datasets, facilitating model development for this complex task.
\end{itemize}

\section{Related Work}
\subsection{fMRI-to-Image Reconstruction}
Benefiting from large-scale datasets like the Natural Scenes Dataset (NSD) \cite{scotti2023reconstructing}, generative vision models conditioned on fMRI signals have demonstrated unprecedented performance in reconstructing static images from brain responses. Existing research primarily focuses on enhancing reconstruction fidelity through improved semantic alignment, such as contrastive learning techniques that align fMRI embeddings with image or text representations \cite{xia2024dream}, or by incorporating low-level image features to preserve visual detail consistency \cite{wang2024mindbridge}. Additional efforts have developed subject-unified methods to address cross-subject alignment and model generalization \cite{scotti2024mindeye2}. Despite significant progress, reconstructing dynamic video sequences presents substantially greater challenges than static images.

\subsection{fMRI-to-Video Reconstruction}
As a pioneering work of fMRI-to-video reconstruction, MindVideo \cite{chen2023cinematic} achieves notable fidelity by aligning fMRI features to CLIP \cite{radford2021learning} space for latent diffusion model prompting. Subsequent studies enhance temporal modeling in fMRI encoders \cite{sun2025neuralflix} or explore cross-subject alignment via fMRI projection \cite{li2024enhancing}. Crucially, most of existing methods are confined to single-shot scenarios, neglecting the multi-shot dynamics inherent in real-world cognition. While NeuroClips \cite{gong2024neuroclips} generates multiple shots by fusing semantically similar keyframes, it relies on post-hoc processing rather than intrinsic fMRI signal decomposition, failing to optimize encoders for disentangling mixed shot information within fMRI windows. Moreover, contrastive alignment in video reconstruction may be challenging due to the temporal resolution mismatch between fMRI and video. In contrast to prior work, we propose to explore the multi-shot video reconstruction by shot-specific fMRI segmentation and keyframe caption decoding for semantically precise reconstruction, circumventing contrastive alignment constraints.

\section{Method}
\begin{figure*}[htbp]
\centering
\includegraphics[width=\textwidth]{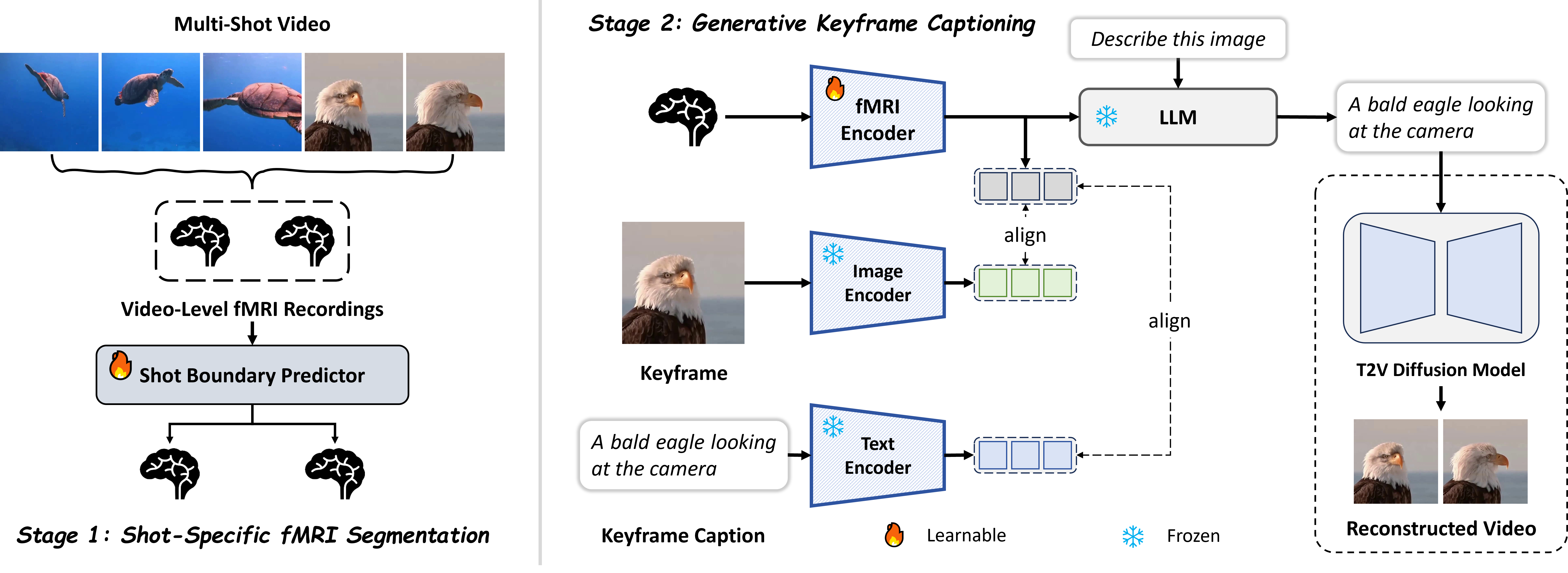}
\caption{Overview of our proposed method.}
\label{fig:Method}
\end{figure*}

Our method can be divided into three main stages, as shown in Figure~\ref{fig:Method}. In the first stage, the shot boundary predictor partitions fMRI into shot-specific intervals. Each segmented fMRI is then decoded to shot-specific keyframe caption via direct interaction with a LLM. These captions serve as precise semantic prompts input to a text-to-video diffusion model for final video synthesis.

\subsection{Dataset Synthesis}
Given the absence of publicly available multi-shot fMRI-video datasets, we develop novel synthesis strategies to enable effective model training. Our approach leverages two public resources: CC2017 \cite{wen2018neural} and Li et al.'s fMRI-WebVid dataset \cite{li2024enhancing}.

\subsubsection{Synthesis from fMRI-WebVid} Using 2,000 single-shot training clips and 400 test clips from fMRI-WebVid (each 4s video with 5 fMRI scans), we generate multi-shot videos by randomly concatenating distinct clips while preserving 4-second duration. To enhance data diversity, we vary shot duration ratios across samples, including partial-shot allocations (e.g., 2:3 fMRI frame split) and degenerate cases reducing to single-shot. For each resulting shot, we extract the middle frame of its video segment as shot-specific keyframes and generate captions using BLIP-2 \cite{li2023blip}. This yields 20,000 training and 1,000 test samples (fMRI-WebVid-Syn) with aligned fMRI sequences, video keyframes, and keyframe captions.

\subsubsection{Synthesis from CC2017} From 1,440 training and 400 test clips in CC2017 (each 6s video with 3 fMRI scans), we synthesize two-shot videos by concatenating distinct clips while maintaining 6-second duration. The number of synthesized fMRI scans is set to 4, considering about the duration of each shot. To ensure single-shot sources, we first apply SceneSeg \cite{rao2020local} to decompose original videos into constituent shots at scene boundaries. For each synthesized video, we randomly select two shots from different source clips, temporally cropping each according to sampled duration ratios (e.g., 3:1) before concatenation. Following the fMRI-WebVid-Syn protocol for keyframe extraction and caption generation, we produce 20,000 training and 1,000 test samples (CC2017-Syn) with aligned fMRI sequences, video keyframes, and keyframe captions.

\subsection{Shot-Specific fMRI Segmentation}
A primary challenge in multi-shot video reconstruction is the temporal mixing of neural signals across different shots. Achieving semantically precise reconstruction thus requires decomposing video-level fMRI into shot-specific components. The most intuitive approach is to detect shot boundaries, transforming the problem into sequence boundary prediction for subsequent fMRI separation and aggregation. Therefore, we propose to detect boundaries by introducing a shot boundary predictor.

Leveraging the fMRI encoder from \cite{li2024enhancing}, we get fMRI embeddings $emb_f\in \mathbb{R}^{M\times c}$, where $M$ is the number of fMRI scans, and $c=1024$ represents the embedding dimension. Theses embeddings are then processed by our proposed shot boundary predictor, which comprises a two-layer bidirectional LSTM (Bi-LSTM) to model bidirectional temporal dependencies in fMRI signals, and a linear layer generating boundary probabilities.

Formally, given fMRI embeddings $emb_f\in \mathbb{R}^{M\times c}$, the boundary probabilities are computed as:
\begin{equation}
H = \text{Bi-LSTM}(emb_f)
\end{equation}
\begin{equation}
P = WH + b
\end{equation}
where $H \in \mathbb{R}^{M \times d}$ are hidden states ($d = 512$), $P = [p_1, p_2, \ldots, p_{M-1}]$ denotes boundary probabilities and $p_i$ represents the boundary probability of a boundary between fMRI scans $i$ and $i + 1$.

The model is optimized via binary cross-entropy loss:
\begin{equation}
\mathcal{L}_{sbp} = -\frac{1}{M-1} \sum_{i=1}^{M-1} [y_i \log p_i + (1-y_i) \log(1-p_i)]
\end{equation}
where $y_i \in \{0,1\}$ indicates ground-truth boundaries, and the true number of shots $N$ satisfies
\begin{equation}
N = 1 + \sum_{i=1}^{M-1} y_i
\end{equation}

At inference, by binarizing $p_i$ with a threshold $\tau$, we get
\begin{equation}
o_i = \begin{cases} 
1 & \text{if } p_i > \tau \\
0 & \text{otherwise}
\end{cases}
\end{equation}
where $\tau = 0.5$ in this work. Using the binarized boundaries $o_i$, we partition the fMRI sequence into $\tilde{N}$ segments, where $\tilde{N} = 1 + \sum_{i=1}^{M-1} o_i$ is the predicted number of shots. The shot-specific embeddings are then aggregated as $emb_f^s \in \mathbb{R}^{\tilde{N} \times c}$.

\subsection{Generative Keyframe Captioning}
Beyond temporal signal mixing, the fMRI-video temporal resolution mismatch makes direct reconstruction via contrastive alignment challenging. However, human cognition encodes experiences through semantic abstractions of key events rather than continuous visual streams. We therefore reformulate the task as keyframe-centered semantic reconstruction, where decoding keyframe captions bypasses strict temporal alignment requirements. Specifically, we learn to generate keyframe captions directly from shot-specific fMRI signals using an LLM, and the keyframe captions are then used for final video synthesis.

Using ground-truth shot boundaries during training, we obtain shot-specific fMRI embeddings $emb_f^s\in \mathbb{R}^{N×c}$ and concatenate them with an instruction prompt and input into a frozen LLM. Leveraging the multimodal understanding capabilities of the LLM, the dialogue format is structured as follows: 
\begin{flushleft}
System: [system message] \\
User: $<$~instruction~$>$ $<$~fMRI embedding~$>$ \\
Assistant: $<$~answer~$>$
\end{flushleft}
here, the tag $<$~instruction~$>$ denotes natural language query, while $<$~image~$>$ is a placeholder for fMRI embedding. The model generates the response $<$~answer~$>$ as predicted captions. The objective for optimizing this decoding process is to minimize the text modeling loss $\mathcal{L}_{caption}$, which evaluates the ability of LLM to generate target captions from fMRI embeddings. This loss is formally defined as the negative log-likelihood of the target captions given context:

\begin{equation}
\mathcal{L}_{caption} = -\sum_{k=1}^{T} \log P_\theta \left( t_k | t_{<k}, I; emb_f^s \right)
\end{equation}
where $T$ is the length of target text, $t_k$ is the $k$-th token, $t_{<k}$ represents the preceding tokens, $I$ is the input prompt ('Describe this image $<$~image~$>$' in this work), and $P_\theta$ is the token probability distribution parameterized by LLM weights $\theta$.

We empirically found that introducing contrastive alignment and noise prediction during training can improve the final results. Given \{keyframe, keyframe caption\} pairs, CLIP loss is calculated for fMRI-keyframe and fMRI-caption pairs. With fixed CLIP encoders, we obtain keyframe embedding $emb_i \in \mathbb{R}^{N \times c}$ and text embedding $emb_t \in \mathbb{R}^{N \times c}$. The contrastive alignment loss is:

\begin{equation}
\mathcal{L}_{align} = \frac{1}{2} \left( \mathcal{L}_{CLIP}(emb_f^s, emb_i) + \mathcal{L}_{CLIP}(emb_f^s, emb_t) \right)
\end{equation}

We also freeze the U-Net of video diffusion model for noise prediction, with MSE loss:

\begin{equation}
\mathcal{L}_{mse} = \mathbb{E}_{emb_f^s, \epsilon_{gt}^t \sim \mathcal{N}(0,1), t} \left[ \left\| \epsilon_{gt}^t - \epsilon_{pr}^t \right\|_2^2 \right]
\end{equation}
where $\epsilon_{pr}^t = \text{U-Net}(emb_i^t, emb_f^s, t)$ is predicted noise conditioned on shot-specific fMRI, and $\epsilon_{gt}^t$ is ground-truth noise.

The overall training loss combines all components:

\begin{equation}
\mathcal{L} = \mathcal{L}_{sbp} + \lambda_1 \mathcal{L}_{caption} + \lambda_2 \mathcal{L}_{align} + \lambda_3 \mathcal{L}_{mse}
\end{equation}
where $\lambda_1$, $\lambda_2$, and $\lambda_3$ are learnable parameters for automatic optimization. Only the fMRI encoder and shot boundary predictor are trained while other modules remain frozen.

\section{Experimental Setting}

\subsubsection{Dataset} We evaluated our method on both synthesized and original datasets, including CC2017 and fMRI-WebVid. CC2017 \cite{chen2023cinematic} contains three subjects with fMRI frames acquired using a 3T scanner (TR=2s), where each sample includes a 6s video and 3 fMRI scans. fMRI-WebVid \cite{li2024enhancing} involves five subjects with fMRI data acquired using a 3T scanner sampled at 1 frame per 0.8s. Stimuli videos (596×336) are sourced from WebVid \cite{bain2021frozen}, with each sample containing a 4s video and 5 fMRI scans. For synthesized datasets, we balanced samples across different duration ratios to avoid data bias. CC2017-Syn used fMRI ratios of [(1,3), (2,2), (3,1)] for 4 synthesized fMRI scans, while fMRI-WebVid-Syn used ratios of [(0,5), (2,3), (3,2)]. Synthesized training data originated only from original training data with no test overlap.

\subsubsection{fMRI Preprocessing} Following \cite{qian2023fmri}, each fMRI scan was projected to 32k\_fs\_LR brain surface space through anatomical structure and transformed to a 256×256 single-channel image, where only early and higher cortical regions retained values. fMRI data were averaged across multiple runs for the same video in both datasets.

\subsubsection{Evaluation Metrics} For video reconstruction, we utilized N-way top-K accuracy for semantic evaluation and SSIM for pixel-level assessment. Shot-specific fMRI segmentation employed segmentation accuracy, normalized mutual information (NMI), and adjusted rand index (ARI) following video scene segmentation research \cite{mahon2024hard}. For evaluating LLM-decoded captions, we used the CLIP text score to measure semantic alignment between generated and ground-truth descriptions.

\subsubsection{Implementation Details} For original CC2017, we processed 3 fMRI scans to generate 6s videos at 3 FPS. CC2017-Syn used 4 fMRI scans for 6s/6 FPS output. Original fMRI-WebVid processed 5 fMRI scans into 4s/3 FPS videos, while its synthesized counterpart used 5 fMRI scans for 4s/6 FPS reconstruction. All videos were generated at dimensions of 576×320. Theoretically, any text-to-video diffusion model can be used for video generation based on the decode captions. In this work, ModelScopeT2V \cite{wang2023modelscope} was used as our video generator, performing inference with 30 DDIM steps and adopt a 6.0 classifier-free guidance score. The image encoder and text encoder were initialized using CLIP ViT-H/14 from OpenCLIP \cite{cherti2023reproducible}, and Qwen3-0.6B \cite{yang2025qwen3} served as the LLM decoder.

\section{Experimental Results}
\subsection{Comparison Results}
\begin{figure*}[htbp]
\centering
\includegraphics[width=\textwidth]{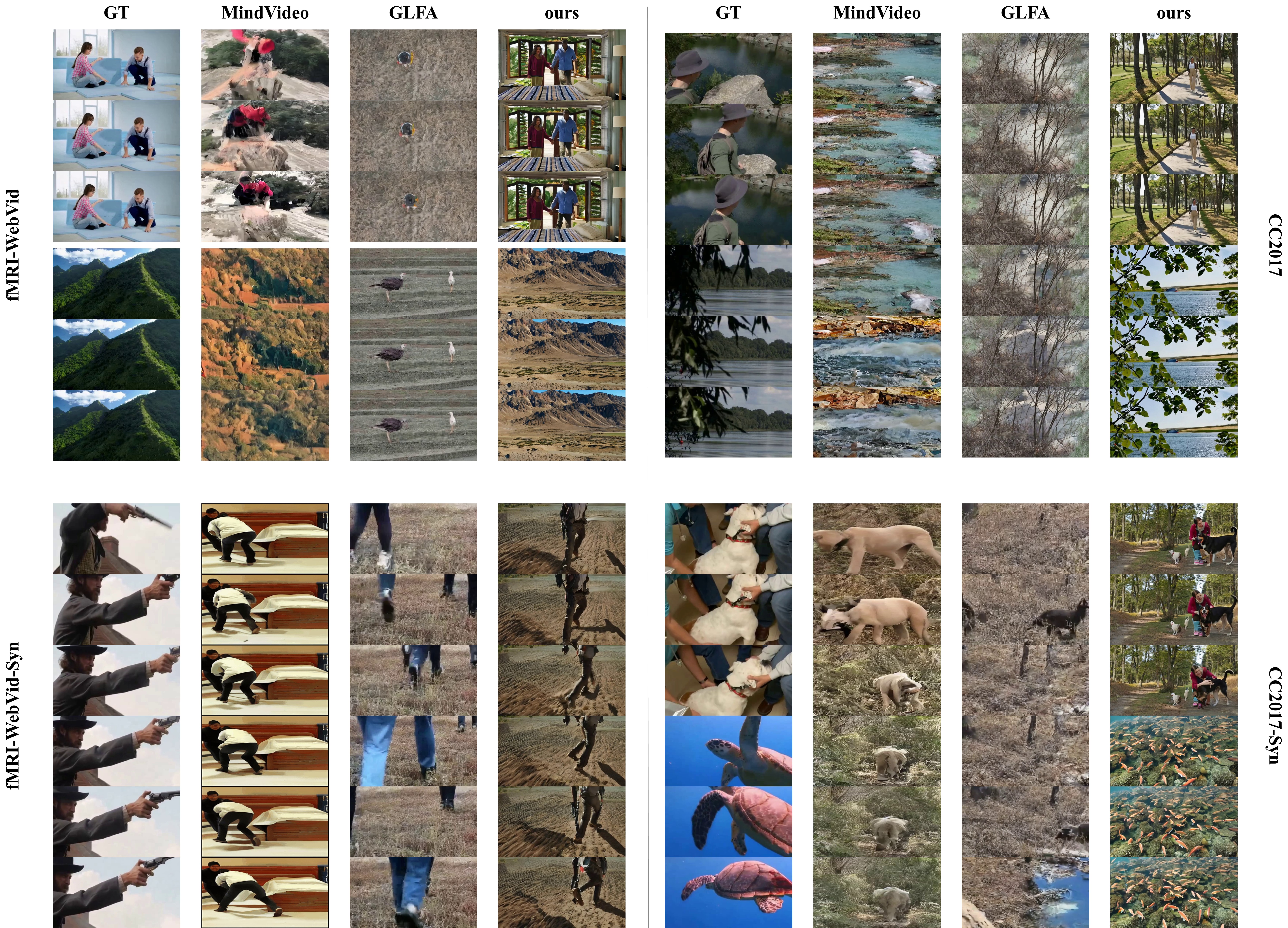}
\caption{Qualitative comparison of fMRI-to-video reconstruction results on four datasets. Visual examples from four datasets demonstrate the superiority of our method compared to three baselines.}
\label{fig:Qualitative results}
\end{figure*}

We compare our method against three fMRI-to-video baselines: MindVideo \cite{chen2023cinematic}, NeuroClips \cite{gong2024neuroclips}, and GLFA \cite{li2024enhancing}. Visual comparisons are shown in Figure~\ref{fig:Qualitative results}, and quantitative results are presented in Table~\ref{tab:comparison results}.

\begin{table*}[htbp]
\centering
\begin{tabular}{ccccccc}
\toprule
\multirow{3}{*}{Dataset} & \multirow{3}{*}{Model} & \multicolumn{4}{c}{Video-Based} & Frame-Based \\
\cmidrule(lr){3-6} \cmidrule(lr){7-7}
& & \multicolumn{2}{c}{Semantic-Level} & \multicolumn{2}{c}{Semantic-Level} & Pixel-Level \\
\cmidrule(lr){3-4} \cmidrule(lr){5-6} \cmidrule(lr){7-7}
& & 2-way$\uparrow$ & 50-way$\uparrow$ & 2-way$\uparrow$ & 50-way$\uparrow$ & SSIM$\uparrow$ \\
\midrule
\multirow{3}{*}{fMRI-WebVid} & MindVideo & 0.736±0.04 & 0.075±0.01 & 0.760±0.03 & 0.109±0.01 & 0.097 \\
& GLFA & 0.790±0.03 & 0.107±0.01 & 0.729±0.03 & 0.118±0.01 & 0.143 \\
& \textbf{ours} & \textbf{0.790±0.03} & \textbf{0.135±0.01} & \textbf{0.817±0.03} & \textbf{0.183±0.02} & \textbf{0.145} \\
\midrule
\multirow{3}{*}{fMRI-WebVid-Syn} & MindVideo & 0.788±0.03 & 0.117±0.01 & 0.735±0.03 & 0.122±0.01 & 0.095 \\
& GLFA & 0.800±0.03 & 0.109±0.01 & 0.727±0.04 & 0.092±0.01 & 0.108 \\
& \textbf{ours} & \textbf{0.819±0.03} & \textbf{0.122±0.01} & \textbf{0.803±0.03} & \textbf{0.138±0.01} & \textbf{0.129} \\
\midrule
\multirow{4}{*}{CC2017} & MindVideo & 0.853±0.03 & 0.202±0.02 & 0.792±0.03 & 0.172±0.01 & 0.171 \\
& NeuroClips & 0.834±0.03 & 0.220±0.01 & \textbf{0.806±0.03} & 0.203±0.01 & 0.211 \\
& GLFA & 0.871±0.03 & 0.219±0.02 & 0.715±0.04 & 0.096±0.01 & 0.083 \\
& \textbf{ours} & \textbf{0.891±0.03} & \textbf{0.235±0.02} & 0.800±0.03 & \textbf{0.206±0.01} & \textbf{0.244} \\
\midrule
\multirow{3}{*}{CC2017-Syn} & MindVideo & 0.813±0.03 & 0.164±0.01 & 0.780±0.03 & 0.107±0.01 & 0.107 \\
& GLFA & 0.877±0.02 & 0.181±0.02 & 0.752±0.04 & 0.087±0.01 & 0.124 \\
& \textbf{ours} & \textbf{0.889±0.02} & \textbf{0.235±0.02} & \textbf{0.781±0.03} & \textbf{0.140±0.01} & \textbf{0.196} \\
\bottomrule
\end{tabular}
\caption{Quantitative comparison of fMRI-to-video reconstruction methods across four datasets, including two original datasets (fMRI-WebVid and CC2017) and two synthesized datasets (fMRI-WebVid-Syn and CC2017-Syn).}
\label{tab:comparison results}
\end{table*}

According to Table~\ref{tab:comparison results}, our method outperforms all baselines, particularly in semantic-level metrics, demonstrating the effectiveness of our approach. Specifically, on the original fMRI-WebVid dataset, our method achieves a 7.5$\%$ improvement in frame-based 2-way classification score compared to the best baseline, while the 50-way classification score shows a substantial 67.9$\%$ improvement. These results suggest that decoding keyframe captions provides a more effective solution for fMRI-to-video reconstruction.

The visual comparisons in Figure~\ref{fig:Qualitative results} reveal that our shot-specific fMRI division strategy significantly contributes to multi-shot video reconstruction quality. In contrast, other baselines exhibit obvious quality degradation and fail to effectively reconstruct coherent multi-shot sequences.

\subsection{Ablation Results}
\subsubsection{Impact of Shot Segmentation} To evaluate the effectiveness of our proposed shot-specific fMRI segmentation, we conducted ablation studies focusing on three key aspects: (1) segmentation performance, semantic caption decoding accuracy, and final video reconstruction quality. We compare our full method incorporating the Shot Boundary Predictor (w/ $\mathcal{L}_{sbp}$) against a baseline (w/o $\mathcal{L}_{sbp}$) where $\mathcal{L}_{sbp}$ loss component is disabled. In the w/o $\mathcal{L}_{sbp}$ baseline, the entire fMRI sequence is processed as a single unit and the fMRI encoder is optimized using a weighted sum of the lossess $\mathcal{L}_{caption}$, $\mathcal{L}_{align}$, and $\mathcal{L}_{mse}$. Here, $\mathcal{L}_{caption}$ is trained to decode a single, video-level caption describing the entire multi-shot sequence.

\begin{table*}[htbp]
\centering
\begin{tabular}{ccccccc}
\toprule
\multirow{2}{*}{Shot Segmentation} & \multirow{2}{*}{Caption CLIP} & \multicolumn{3}{c}{Segmentation Metrics} & \multicolumn{2}{c}{Video Reconstruction Metrics} \\
\cmidrule(lr){3-5} \cmidrule(lr){6-7}
& & ACC$\uparrow$ & ARI$\uparrow$ & NMI$\uparrow$ & 2-way$\uparrow$ & 50-way$\uparrow$ \\
\midrule
w/o $\mathcal{L}_{sbp}$ & 0.177 & - & - & - & 0.814±0.03 & 0.112±0.01 \\
w/ $\mathcal{L}_{sbp}$ & 0.304 & 0.685 & 0.683 & 0.690 & 0.819±0.03 & 0.122±0.01 \\
\bottomrule
\end{tabular}
\caption{Ablation results of shot-specific fMRI segmentation on fMRI-WebVid-Syn dataset.}
\label{tab:ablation}
\end{table*}

\begin{table*}[htbp]
\centering
\begin{tabular}{cccccc}
\toprule
\multirow{3}{*}{Prompt} & \multicolumn{4}{c}{Video-Based} & Frame-Based \\
\cmidrule(lr){2-5} \cmidrule(lr){6-6}
& \multicolumn{2}{c}{Semantic-Level} & \multicolumn{2}{c}{Semantic-Level} & Pixel-Level \\
\cmidrule(lr){2-3} \cmidrule(lr){4-5} \cmidrule(lr){6-6}
& 2-way$\uparrow$ & 50-way$\uparrow$ & 2-way$\uparrow$ & 50-way$\uparrow$ & SSIM$\uparrow$ \\
\midrule
fMRI Only & 0.810±0.03 & 0.097±0.01 & 0.790±0.03 & 0.130±0.01 & \textbf{0.145} \\
Text Only & \textbf{0.822±0.03} & \textbf{0.147±0.013} & 0.815±0.03 & \textbf{0.181±0.01} & 0.144 \\
Dual-Modal & 0.809±0.03 & 0.108±0.01 & \textbf{0.821±0.03} & 0.171±0.02 & 0.101 \\
\bottomrule
\end{tabular}
\caption{Ablation results of different prompt settings for video diffusion model.}
\label{tab:prompt_ablation}
\end{table*}

As shown in Table~\ref{tab:ablation}, the shot boundary predictor achieves a segmentation accuracy of 0.685, and scores of 0.683 in ARI, and 0.690 in NMI, demonstrating the capability of shot boundary predictor to effectively identify transitions between distinct visual shots within the fMRI signal.

\begin{figure}[htbp]
\centering
\includegraphics[width=0.5\textwidth]{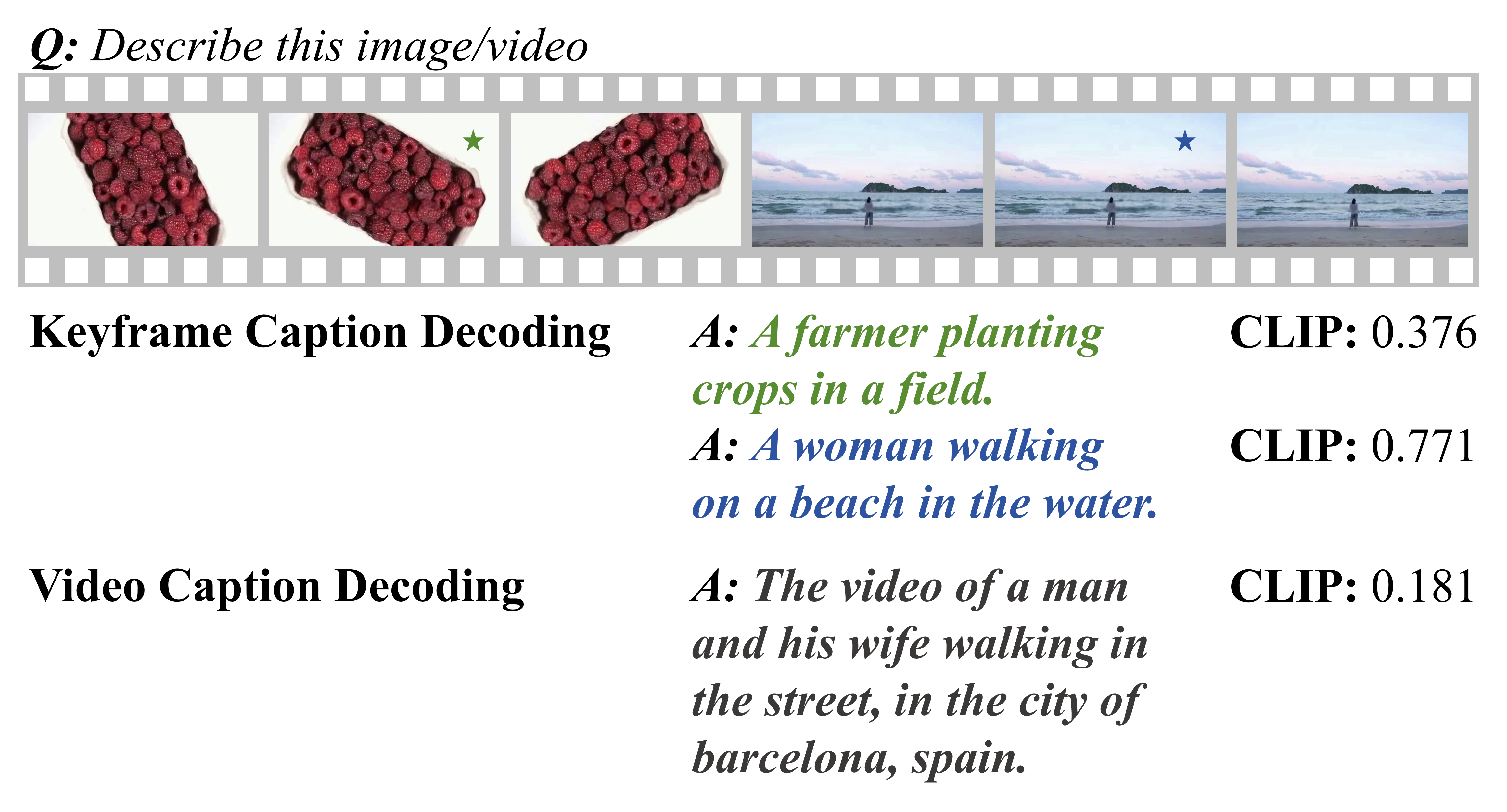}
\caption{Comparison of decoded captions with and without shot-specific fMRI segmentation. }
\label{fig:Caption Decoding Examples}
\end{figure}

We also evaluate the impact of segmentation on the semantic precision of decoded captions. For the w/o $\mathcal{L}_{sbp}$ baseline, the decoded caption represents the entire video. Therefore, CLIP similarity is calculated against the ground-truth video-level caption. In contrast, for w/ $\mathcal{L}_{sbp}$ method, CLIP similarity is computed between the ground-truth keyframe captions and decoded keyframe captions for each individual shot. Results in Table~\ref{tab:ablation} show that introducing shot segmentation improves CLIP similarity by 71.8$\%$. This substantial improvement validates that dividing fMRI signals into shot-specific components and decoding keyframe captions per shot yields more semantically precise descriptions than attempting to decode a single, aggregated video-level caption from the mixed fMRI signal. Qualitative examples in Figure~\ref{fig:Caption Decoding Examples} illustrate that captions from w/o $\mathcal{L}_{sbp}$ are often semantically imprecise and biased towards unrelated content instead of reconstructing any specific shot, whereas captions from w/ $\mathcal{L}_{sbp}$ accurately focus on the core content of each individual shot.

Finally, we assess the impact of segmentation on the ultimate video reconstruction. Quantitative metrics in Table~\ref{tab:ablation} show improvements of w/ $\mathcal{L}_{sbp}$ compared to w/o $\mathcal{L}_{sbp}$. This demonstrates that the enhanced semantic precision achieved through shot segmentation and keyframe caption decoding also contributes to higher-fidelity dynamic video reconstructions.

In summary, the divide-and-decode strategy effectively mitigates the semantic ambiguity inherent in processing mixed fMRI signals from multi-shot sequences, enabling the final high-fidelity multi-shot video reconstruction.

\subsubsection{Impact of LLM Decoding} To evaluate the effectiveness of caption decoding against contrastive alignment for semantic extraction from fMRI, we conduct ablation experiments evaluating CLIP similarity under different training regimes. For contrastive alignment, since there are no decoded captions, we calculate the CLIP similarity between fMRI embeddings and ground-truth caption embeddings.

\begin{table}[htbp]
\centering
\begin{tabular}{cccc}
\toprule
\multicolumn{3}{c}{Loss Function} & Metric \\
\cmidrule(lr){1-3} \cmidrule(lr){4-4}
$\mathcal{L}_{caption}$ & $\mathcal{L}_{align}$ & $\mathcal{L}_{mse}$ & CLIP$\uparrow$ \\
\midrule
- & $\checkmark$ & - & 0.283 \\
- & $\checkmark$ & $\checkmark$ & 0.280 \\
$\checkmark$ & - & - & 0.302 \\
$\checkmark$ & $\checkmark$ & - & 0.313 \\
$\checkmark$ & - & $\checkmark$ & 0.300 \\
$\checkmark$ & $\checkmark$ & $\checkmark$ & \textbf{0.336} \\
\bottomrule
\end{tabular}
\caption{Ablation results on semantics extraction methods on fMRI-WebVid-Syn dataset.}
\label{tab:semantics_ablation}
\end{table}

As shown in Table~\ref{tab:semantics_ablation}, caption decoding ($\mathcal{L}_{caption}$ only) improves CLIP similarity by 6.7$\%$ over the alignment baseline ($\mathcal{L}_{align}$ only), demonstrating that decoding text descriptions better reconstruct semantics by mitigating temporal ambiguity. Although caption decoding outperforms the alignment baseline, semantic extraction is further enhanced by the multi-task framework that combines alignment, decoding, and reconstruction objectives. We ascribe this to the complementary information provided by different tasks, where alignment task helps preserve structural details while decoding primarily captures semantics.

\subsubsection{Impact of Prompt Settings} Our method uses decoded keyframe captions as input prompts for the video generation model. To validate this design choice, we compare three prompt configurations: fMRI-only, text-only, and dual-modal. The fMRI-only setting uses fMRI embeddings directly as prompt embeddings for video diffusion model, while the dual-modal approach combines fMRI embeddings and text embeddings of decoded captions with equal weighting. As shown in Table~\ref{tab:prompt_ablation}, using only decoded keyframe captions as prompts achieves the best results, particularly on semantic-level metrics. Notably, dual-modal results are worse than text-only and even underperform fMRI-only on some metrics. We ascribe this degradation to the fact that combining embeddings may alter their representations in the latent space, contrary to our expectation that the combination would preserve visual details from fMRI while maintaining high semantic quality from decoded captions.

\section{Conclusion}
In this work, we propose a novel divide-and-decode framework for reconstructing multi-shot videos from fMRI with high semantic fidelity. The shot-specific fMRI segmentation explicitly decouples mixed neural signals, providing cleaner shot-specific components for later semantics extraction and final video reconstruction. Decoding keyframe captions from shot-specific fMRI mitigates the temporal ambiguity caused by fMRI-video temporal resolution mismatch. By integrating these innovations, our framework achieves high-fidelity multi-shot reconstruction where prior methods fail. As one of the pioneering explorations in multi-shot fMRI decoding, we hope that our method can inspire future multi-shot video reconstruction endeavors.

\bibliography{aaai2026}


\end{document}